\documentclass[11pt,a4paper]{article}
\usepackage{times,latexsym}
\usepackage{url}
\usepackage[T1]{fontenc}

\usepackage[acceptedWithA]{tacl2021v1}
\usepackage{xspace,mfirstuc,tabulary}

\newif\iftaclinstructions
\taclinstructionsfalse % AUTHORS: do NOT set this to true
\iftaclinstructions
\renewcommand{\confidential}{}
\renewcommand{\anonsubtext}{(No author info supplied here, for consistency with
TACL-submission anonymization requirements)}
\newcommand{\instr}
\fi

\iftaclpubformat % this "if" is set by the choice of options

\else

\fi

\usepackage{tcolorbox}
\usepackage{algorithm}
\usepackage{algpseudocode}

\usepackage{amsmath}
\usepackage{graphicx}
\usepackage{tabularx}
\usepackage{xspace}
\usepackage[obeyfinal]{todo}
\usepackage{stfloats}
\usepackage[group-separator={,},group-minimum-digits={3}]{siunitx}
\usepackage{CJK}
\usepackage{booktabs}

\newcommand{\mgen}{\ensuremath{M_{gen}}} % generator model
\newcommand{\mopt}{\ensuremath{M_{opt}}} % optimizer model

\algrenewcommand\algorithmicindent{1em}

\title{SEO: Stochastic Experience Optimization for Large Language Models}

\author{Jitao Xu$^{1,2*}$ \quad Hongyun Zhou$^{3*}$ \quad Lei Shen$^1$\Thanks{Equal contribution.} \quad Conghui Zhu$^3$ \quad \vspace{1mm}\\
\textbf{Jin Huang}$^1$ \quad \textbf{Yitao Duan}$^{1}$\Thanks{Corresponding authors.}\quad \vspace{1mm}\\
$^1$NetEase Youdao, Beijing, China\\
$^2$Department of Computer Science and Technology, Tsinghua University, Beijing, China\\
$^3$Faculty of Computing, Harbin Institute of Technology \\
{\tt \{xujt01,duan\}@rd.netease.com}
}

\date{}
\begin{document}

\maketitle
\begin{abstract}
Large Language Models (LLMs) can benefit from useful experiences to improve their performance on specific tasks. However, finding helpful experiences for different LLMs is not obvious, since it is unclear what experiences suit specific LLMs. Previous studies intended to automatically find useful experiences using LLMs, while it is difficult to ensure the effectiveness of the obtained experience. In this paper, we propose Stochastic Experience Optimization (SEO), an iterative approach that finds optimized model-specific experience without modifying model parameters through experience update in natural language. In SEO, we propose a stochastic validation method to ensure the update direction of experience, avoiding unavailing updates. Experimental results on three tasks for three LLMs demonstrate that experiences optimized by SEO can achieve consistently improved performance. Further analysis indicates that SEO-optimized experience can generalize to out-of-distribution data, boosting the performance of LLMs on similar tasks.
\end{abstract}

\section{Introduction\label{sec:intro}}

Large Language Models (LLMs) have demonstrated impressive abilities in various tasks \citep{Brown20gpt3,Chowdhery22palm,Ouyang22training,Touvron23llama}. However, the performance of LLMs are still not perfect on specific tasks \citep{Kocon23chatgpt}. Though fine-tuning LLMs is an effective approach to improve their performance on specific tasks, it requires expensive computational resources and is even impossible when LLMs are closed sourced or only accessible via API like GPT-4 \citep{Openai23gpt4}.

Recent works have shown that LLMs can benefit from useful experiences to improve their performance at inference time without updating their parameters \citep{Dalvi22towards,Madaan22memory,Shinn23reflexion}. Experiences filled into prompts are often presented as lists of general insights \citep{Majumder23clin} or rules \citep{Yang23failures,Zhao23expel} in natural language, providing helpful guidance on the execution of specific tasks. Nevertheless, finding useful experiences for LLMs is not an easy task. Intuitively, handcrafted experiences are preferred \citep{Dalvi22towards,Zhao23expel}. However, they are not guaranteed to be optimal for various LLMs \citep{Zhao23expel,Zhao23selfexplain} and may even lead to performance drop in some cases (as we show in Section~\ref{ssec:init}). Useful experiences are often model-specific \citep{Chen23mapo,Yang24optimizers}, and it is hard for human to craft experiences for various domains and different LLMs. 

Some studies have also made efforts to find useful experiences automatically via LLMs. \citet{Zhao23expel,Majumder23clin} show that LLMs can iteratively generate and refine experiences by self-reflecting \citep{Shinn23reflexion} over their generation history of similar tasks. Nevertheless, the experience refinement relies solely on human-designed prompts that instruct LLMs to generate nominally useful experiences without explicit training labels or updating direction, which does not guarantee an effective optimization. On the other hand, some works have utilized LLMs as an optimizer \citep{Yang24optimizers} to find concise instructions in a natural language optimization process, demonstrated to be effective \citep{Fernando23promptbreeder,Zhou23human,Pryzant23automatic,Tang24unleashing}. These studies are inspiring as they find instructions that yield improved performance.

\begin{figure*}[ht]
  \center
  \includegraphics[width=.95\textwidth]{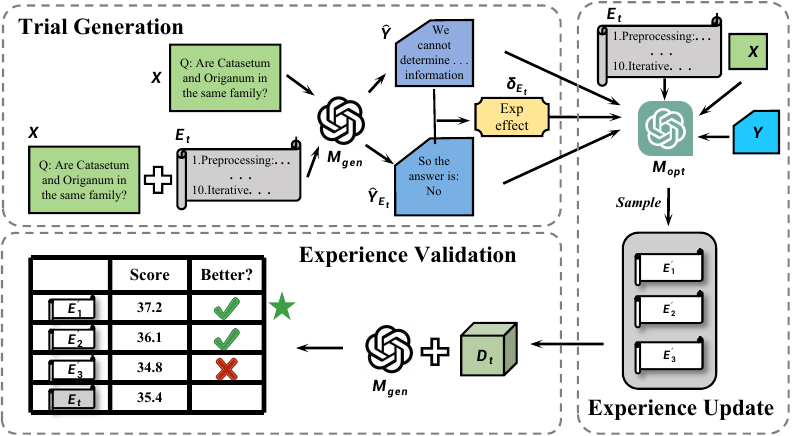}
  \caption{Overview of a training step of the SEO process. We first gather trials with and without experience ($\hat{Y}_{E_t}$ and $\hat{Y}$, respectively) using a generator model \mgen{} and compute experience effect $\delta_{E_t}$. We then use an optimizer model \mopt{} to sample refined candidate experiences $E'_i$ by taking the training example $X$ and $Y$, the trials $\hat{Y}_{E_t}$ and $\hat{Y}$, the experience $E_t$ and the effect of experience $\delta_{E_t}$ as input. Candidate experiences $E'_i$ are then validated on a stochastic set $D_t$ to select the best experience that surpasses the current experience $E_t$ for next iteration.\label{fig:seo}}
\end{figure*}

In this paper, we propose an automatic and effective experience optimization framework, named Stochastic Experience Optimization (SEO). It is capable of finding useful and model-specific experiences without modifying the parameters of LLMs. SEO is conducted through a training process completely in natural language. At each training step, we apply current experience to a generator LLM to collect execution trials. An optimizer LLM then uses the trials and modifies the current experience. Experience is validated through a performance evaluation step using a stochastic validation set to provide a clear optimization direction. With the validation step, SEO achieves generally effective training steps. 

We investigate our proposed SEO method on three tasks, including multi-hop Question Answering (QA), Machine Translation (MT) and Text Classification over seven datasets using three LLMs. Experiences obtained by SEO consistently improve the performance of various LLMs across almost all tasks. Our main contributions can be summarized as follows: 
\begin{itemize}
  \item We propose an automatic experience optimization framework SEO that is able to find useful and model-specific experience without modifying its parameters. SEO is more effective owing to the explicit optimization direction compared to prior works. 
  \item SEO is a stable and general framework that is model-agnostic and task-agnostic. The obtained experiences generally boost the performance on various tasks for various LLMs.
  \item  Experimental analyses showcase that experience optimized by SEO has the potential to generalize to out-of-distribution data under similar tasks.
\end{itemize}

\section{Stochastic Experience Optimization\label{sec:seo}}

SEO is inspired by the training process of machine learning systems, and designed as an automatic experience optimization framework to improve the performance of specific LLMs on specific tasks. The experience is regarded as model parameters in machine learning systems that needs to be optimized via a sort of ``stochastic gradient descent'' (SGD) approach over training examples.

\begin{algorithm}[ht]
  \small
  \caption{SEO\label{algo:seo}}
  \begin{algorithmic}[1]
      \Require \mgen{}, \mopt{}, training data $D$, maximum training step $max\_step$
      \State Initialize  $E_{0}$
      \State $t \leftarrow 0$
      \While {$t < max\_step$}
        \State Batch $\leftarrow$ \{$(X,Y)\}_n$\
        \State $P,S \leftarrow [ ]$
        \For{$(X, Y)$ in Batch}
          \State $\hat{Y} \leftarrow \mgen(X)$ 
          \State $\hat{Y}_{E_t} \leftarrow \mgen(X, E_t)$
          \State $\delta_{E_t} \leftarrow \mathrm{Eval}(\hat{Y}_{E_t}) - \mathrm{Eval}(\hat{Y})$
          \For{$i = 1,\dots,k$}
            \State $E_i \leftarrow$ $\mopt(X, Y, E_t, \hat{Y}, \hat{Y}_{E_t}, \delta_{E_t})$
            \State $P \leftarrow P + E_i$
          \EndFor
        \EndFor
        \State Randomly sample $D_{t} \subset D$
        \For{$E$ in $P$}
            \State $S \leftarrow S + \mathrm{Eval}(\mgen(D_t, E))$
        \EndFor
        \State $S_t \leftarrow \mathrm{Eval}(\mgen(D_t, E_t))$
        \If {$\max(S) > S_t$}
           \State $E_{t+1} \leftarrow \arg\max_{E\in P}S $
        \Else
            \State $E_{t+1} \leftarrow E_{t} $
        \EndIf
        \State $t \leftarrow t + 1$
      \EndWhile
      \State\Return Best $E$ in $E_1,\dots,E_{max\_step}$
  \end{algorithmic}
\end{algorithm}

SEO process, as shown in Figure~\ref{fig:seo}, mainly contains five components including a generator model \mgen, an optimizer model \mopt, the initial experience $E_0$ that needs to be optimized, training data $D$ and stochastic validation sets. We divide SEO into three main steps: Trial Generation, Experience Update and Experience Validation. For a specific task, in Trial Generation, the generator model \mgen{} is used to execute a training sample with and without using the current experience. The generated responses are considered as trials. Experience Update aims to instruct the optimizer model \mopt{} to refine the existing experience with comprehensive information of a comparison between trials with and without experience and also the correct answer, leading to more promising candidate experience. In addition, we use a small stochastic validation set to evaluate the generated candidate experiences and then obtain actual useful experience in the Experience Validation step. This step actually provides a clear update direction like SGD. During training, the generator and optimizer models are both frozen. The optimized experience corresponding to the generator model is the target product of the SEO process. Algorithm~\ref{algo:seo} illustrates the SEO process, which is detailed below.

\subsection{Trial Generation}

SEO requires an initial experience $E_0$ to start the optimization process. $E_0$ is updated iteratively during training. Given an input training example $X$ and an experience $E_t$ at time step $t$, we first use the generator model \mgen{} to produce trials with and without using the experience $\hat{Y}_{E_t} = \mgen(X, E_t)$ and $\hat{Y} = \mgen(X)$, respectively. To give the optimizer model \mopt{} a clearer view of the effectiveness of $E_t$, different from previous work \citep{Majumder23clin,Zhao23expel}, we also evaluate the effect $\delta_{E_t}$ of applying $E_t$ to \mgen{} by calculating the performance change between $\hat{Y}_{E_t}$ and $\hat{Y}$: 
\begin{equation}
  \delta_{E_t} = \mathrm{Eval}(\hat{Y}_{E_t}) - \mathrm{Eval}(\hat{Y}),
\end{equation} 
where $\mathrm{Eval}(\cdot)$ is the evaluation score of a response trial. $\delta_{E_t}$ could be either positive, negative or even zero. The latter indicates that $E_t$ is not effective enough to \mgen, thus needs to be improved. We show in Appendix~\ref{sec:prompt} the prompts used for \mgen{} on different tasks. 

\subsection{Experience Update\label{ssec:exp-update}}

We conduct experience update using the optimizer model \mopt. For natural language optimization, it is difficult to compute an exact ``gradient'' \citep{Tang24unleashing}. To this end, we give the optimizer model \mopt{} more information that may help it to generate better updates. It takes the training example $(X, Y)$, the current experience $E_t$, the generated trials $\hat{Y}$ and $\hat{Y}_{E_t}$, and also the effect of experience $\delta_{E_t}$ as input to produce an updated experience:
\begin{equation}
  E' = \mopt(X, Y, E_t, \hat{Y}, \hat{Y}_{E_t}, \delta_{E_t}). 
\end{equation}
The inputs $X$, $Y$, $\hat{Y}$ and $\hat{Y}_{E_t}$ provide a more complete context of the application scenario of the experience $E_t$. The effect $\delta_{E_t}$ directly indicates if the experience $E_t$ is useful under this scenario. 

The updated experience is expected to improve the performance of \mgen{} over the task, but not on a single example. We specify the requirements in the optimization prompt used to update the experience, as shown in Figure~\ref{fig:opt-prompt} for the MT task as an example. 

We follow prompt optimization methods \citep{Zhou23human,Fernando23promptbreeder,Yang24optimizers} and sample $k$ times using \mopt, resulting in $k$ candidate experiences $P = \{E'_1,\dots,E'_k\}$ for each training example. Note that sampling candidate experiences is a critical step in the SEO process. It not only increases the diversity of experience, exploring more possibilities, but also makes the validation step possible. Otherwise, without the sampling step, only one candidate experience is generated at each step, which is by default kept for the next iteration.

\paragraph{Size of Experience}

As introduced in Section~\ref{sec:intro}, experience consists of a list of rules or insights written in natural language sentences. In our experiments, we limit the number of rules to around ten in the optimization prompt (See Figure~\ref{fig:opt-prompt}). Our preliminary experiments find that without this constraint, the optimizer model tends to continually add similar new rules, resulting in too many sentences that exceed the context window and greatly distract LLMs, leading to catastrophic performance drop. On the contrary, limiting the number of rules to less than ten causes the optimizer model to specify the experience and only generate example-specific content that cannot generalize to other examples. Note that we do not make a hard constraint about the size of experience in the optimization prompt. Instead, the number constraint only helps \mopt{} to avoid divergence during experience update, while the actual number of rules in the optimized experience is decided by the optimizer model \mopt{} itself.  

\paragraph{Mini Batch Optimization}

Similar to SGD, we perform mini batch optimization by processing a batch of training examples together in a training step. For a batch size $n$, we repeat the trial generation and experience update steps $n$ times for $n$ different training examples using the same experience $E_t$. As a result, we obtain $n\times k$ candidate experiences in $P$ that are evaluated together in a single experience validation step. Mini batch optimization is important to keep the updated experience generalized and delivers more stable updates during SEO.

\begin{figure}[!ht]
\small
\begin{tcolorbox}[colback=gray!5!white,colframe=gray!75!black,title=Prompt to generate candidate experience]
  You are an advanced agent. Your task is to adjust existing rules so that a translation agent can apply these rules to generate better translation.\\

  Below is an agent trying to translate an input sentence.\\
  
  Input: \{\texttt{source\_sentence}\}\\
  
  Here are the existing rules:\\
  
  <Rules>\\
  \{\texttt{experience}\}\\
  </Rules>\\
  
  Without these rules, the agent gives the following translation:\\
  
  <Translation without rules>\\
  \{\texttt{answer\_without\_experience}\}\\
  </Translation without rules>\\
  
  After applying these rules, the agent has changed the translation to:\\
  
  <Translation with rules>\\
  \{\texttt{answer\_with\_experience}\}\\
  </Translation with rules>\\
  
  \{\texttt{exp\_effect}\} The reference translation to this sentence is: "\{\texttt{reference}\}".\\
  
  Please modify the existing rules so that the modified rules can directly help an agent generate better translation.\\
  
  You can EDIT existing rules, DELETE unhelpful or redundant rules, or ADD new rules when necessary. You DO NOT NEED TO FOLLOW the format of existing rules. The rules should be GENERAL, CONCISE and EASY TO APPLY, and should be GENERALLY APPLICABLE to other sentences. DO NOT MENTION sentence-specific contents in the rules.\\
  Please keep the rules as simple as possible. Limit their number to around ten.\\
  
  Adjusted rules:
\end{tcolorbox}
\caption{Prompt used for the optimizer model to generate candidate experiences for MT. 
Contents in curly braces are variables.
\label{fig:opt-prompt}}
\end{figure}

\subsection{Experience Validation}

Among all the candidate experiences, we aim to validate and select the most helpful experience as the update $E_{t+1}$ for the next training step. For each candidate experience $E'_i \in P$, we evaluate whether it actually helps \mgen{} to improve its performance. We randomly sample a small set of $m$ training examples to construct a validation set $D_t$. We then evaluate the performance of applying $E_t$ and $E'_i\in P$ to \mgen{} on the validation set $D_t$, computed as:
\begin{equation}
 S_i = \mathrm{Eval}(\mgen(D_t, E'_i)).  
\end{equation}
The obtained scores $S = \{S_1,\dots,S_{nk}\}$ are then compared to the score $S_t$ of applying $E_t$ on the validation set $D_t$. We take the experience that obtains the best performance and also surpasses $S_t$ as a valid experience update and keep the corresponding experience for the next training step $E_{t+1}$. If none of the candidate experiences performs better than $E_t$, we ignore all the candidates and keep $E_t$ for the next training step, which indicates that this training step does not yield a valid experience update.
\begin{equation}
E_{t+1} = 
\begin{cases}
  \arg\max_{E'_i\in P}S&, \mathrm{if} \max(S) > S_t \\
  E_t&, \mathrm{otherwise}\\
\end{cases}
\end{equation}

When optimization forms in natural languages, it is difficult to obtain real gradients like conventional gradient-based optimization \citep{Tang24unleashing}. Therefore, the optimization direction cannot be guaranteed. In SEO, we use the performance scores $S$ on the validation set $D_t$ to simulate the ``descent'' update direction. In other words, the performance improvement on the validation set is regarded as a descent gradient that instructs us to optimize experience. With experience validation, we make sure that an updated experience actually improves the performance of \mgen{} on $D_t$. By sampling $D_t$ at each step, we stochastically improve the quality of experience during the optimization process as SGD. In addition, we use a held-out dev set to select the best experience from optimized experiences of all valid training steps for test.

The validation set $D_t$ is resampled at each training step stochastically. Similar to prompt optimization methods which often sample a subset of training data for evaluation \citep{Pryzant23automatic,Zhou23human}, we use a stochastic validation set to obtain optimized experience for several reasons. First, the experience is relatively long as it contains multiple rules. Therefore, evaluating several candidate experiences on a large amount of examples greatly increases the computational costs during the optimization process, especially when the training set contains thousands of examples. Second, evaluating on a fixed set risks overfitting to the dataset, especially when the size of $D_t$ is relatively small. Prompt optimization approach often consider $50$-$100$ examples as the validation set \citep{Fernando23promptbreeder,Tang24unleashing}, which is unlikely to be general enough if it is a fixed set.

\subsection{Applying Experience}

The optimized experience is integrated into \mgen{} via prompting. We add the following instructions in the generation prompt to let \mgen{} apply these experiences:
{\small
\begin{tcolorbox}[colback=gray!5!white,colframe=gray!75!black,title=Prompt for applying experience]
Here are some rules that might be helpful for you to \{\texttt{task\_description}\}. Try to apply these rules as much as you can.
\\
\\
<Rules>\\
\{\texttt{experience}\}\\
</Rules>
\end{tcolorbox}
}

\section{Experiments\label{sec:exp}}

\subsection{Datasets and Metrics\label{ssec:data}}

We conduct experiments on three distinct tasks that are commonly used natural language processing tasks, namely multi-hop QA, MT and Text Classification. For multi-hop QA, we use HotpotQA \citep{Yang18hotpotqa}. The multi-hop QA task asks models to reason through multiple paragraphs and find the answer of the given question. The dataset contains ten paragraphs with eight distractors and two containing useful information for each question. All paragraphs are used together with the question. We reuse the dev and test split of \citep{Trivedi23interleaving}, in which the dev set contains $100$ examples and the test set contains $500$ examples. Both sets are split from the officially released dev set of \citep{Yang18hotpotqa}. We use the remaining examples in the official dev set as training set in our experiments. We use the Exact Match (EM) score as the evaluation metrics.

For MT, we experiment on two language pairs English-German (En-De) and English-Chinese (En-Zh) in both directions. We use the WMT23 general test set \citep{Kocmi23findings} as our test bed.\footnote{\url{https://www2.statmt.org/wmt23/}} We sample $200$ examples from WMT22 test set \citep{Kocmi22findings} as our held-out dev set, and combine the remaining data with WMT21 test set \citep{Akhbardeh21findings} to build our training data. We use COMET \citep{Rei20comet} with the COMET22 model\footnote{\url{https://huggingface.co/Unbabel/wmt22-comet-da}} \citep{Rei22comet} to evaluate the translation quality.

For Text Classification, we use SST-2 and CoLA from the GLUE benchmark \citep{Wang18glue}. SST-2 is the Stanford Sentiment Treebank which is a sentiment analysis task. CoLA refers to Corpus of Linguistic Acceptability, which is aimed to classify the correctness of a given sentence. We randomly select $200$ examples from the training set as our held-out dev set, and test the performance on the official dev set. The performance is measured by the accuracy score.

\subsection{Experimental Setup\label{ssec:exp}}

For the generator model \mgen, we use \texttt{gpt-3.5-turbo-0613} and Llama-2 models \citep{Touvron23llama} \texttt{Llama-2-7b-chat} and \texttt{Llama-2-13b-chat}. We use a decoding temperature of $0$ for all generator models. For the optimizer model, our preliminary experiments found that using a stronger model leads to better performance, which is inline with \citet{Zhao23expel}. Therefore, we use \texttt{gpt-4-0613} as our optimizer model \mopt{} in all experiments unless otherwise specified. We use a decoding temperature of $0.5$ for the optimizer model. We sample $k = 3$ times in each experience update step. The best performing experience is selected based on the performance on the held-out dev set from optimization process using a batch size of $1$ and $3$. At each training step, we randomly sample $m = 50$ examples from the training set as the stochastic validation set. Due to computational budgets, we use a maximum of $200$ steps for all training process. We will release our code and optimized experiences upon acceptance. 

For the initial experience $E_0$, for each task, we directly ask the optimizer model GPT-4 to generate an experience to perform each task and use the response as the initialization for all generator models as mentioned above. Note that for the MT task, we use the same initialization for all four directions, as we do not specify a specific translation direction when generating the initial experience.

\paragraph{Baselines}

We mainly compare SEO with two baselines: direct answer (Direct) and applying the initial experience (Init), as the main objective of SEO is to find experience that delivers improved performance over Direct and Init. We also compare with CLIN \citep{Majumder23clin} and ExpeL \citep{Zhao23expel}, which are the most relevant methods to our work that also iteratively learn experiences from the trials generated by LLMs, but does not contain an experience validation step. CLIN is designed to complete tasks in the ScienceWorld simulator \citep{Wang22scienceworld}, while ExpeL relies on a ReAct way \citep{Yao23react} to execute tasks. Both methods cannot be directly applied in our experiments. Therefore, we tried our best to reimplement CLIN and ExpeL to suit our tasks. Note that these two approaches only use GPT-3.5 as the generator model. The effect of applying CLIN and ExpeL to other models like Llama-2 models is unknown.

\section{Main Results\label{sec:res}}

We report the overall results in this section for the three tasks we consider with three different LLMs as the generator model. As shown in Tables~\ref{tab:qa}, \ref{tab:mt} and \ref{tab:classification}, in general, SEO is able to find experiences that steadily improves all generator models on various tasks except for one case, suggesting that it is a stable model-agnostic and task-agnostic experience optimization method. On the contrary, CLIN and ExpeL tend to be unstable, as the results on certain tasks and certain models are comparable to our SEO experience, while others lag significantly behind. As stated in Section~\ref{sec:intro}, methods like CLIN and ExpeL cannot guarantee a performance improvement during the optimization process and is therefore not easy to apply as there is no information about what task or model it suits well.
We present detailed results in the following. 

\begin{table}[!ht]
  \center
  \scalebox{0.95}{
  \begin{tabular}{l|ccc}
  \toprule
  Models & GPT-3.5 & Llama-2-13b & Llama-2-7b \\
  \hline
  CLIN   & 40.8 & 20.2 & 10.6 \\
  ExpeL  & 40.2 & 17.2 & \hphantom{0}8.8 \\
  Direct & 37.2 & 19.4 & 19.0 \\
  Init  & 38.6 & 17.8 & \hphantom{0}7.8 \\
  % \hline
  SEO    & \textbf{45.2} & \textbf{29.8} & \textbf{22.6} \\
  \bottomrule
  \end{tabular}
  }
  \caption{Performance of various generator models on HotpotQA measured by EM score. \textit{Direct} refers to direct answer without using experience. \textit{Init} indicates applying the initial experience. \textit{SEO} is obtained by using the best performing experience optimized by SEO.\label{tab:qa}}
\end{table}

\subsection{Multi-Hop Question Answering}

The results for multi-hop QA are shown in Table~\ref{tab:qa}. As it can be seen, the same initial experience does not always improve the performance across different generator models, indicating that useful experience is model-specific. The initial experience boost the performance of GPT-3.5 from $37.2$ to $38.6$, but has a negative effect for Llama-2-13b and even devastates the performance on Llama-2-7b. Note that we use the same initialization for all models. These results indicate that the effectiveness of the experience varies across models and model-specific experience are required to obtain performance improvement.

\begin{table*}[!ht]
  \center
  \scalebox{1}{
  \begin{tabular}{l|ccc|ccc}
  \toprule
  Language & \multicolumn{3}{c|}{En-De} & \multicolumn{3}{c}{De-En} \\
  \hline
  Models & GPT-3.5 & Llama-2-13b & Llama-2-7b & GPT-3.5 & Llama-2-13b & Llama-2-7b \\
  \hline
  CLIN   &  83.98 & 72.16 & 69.53 & 85.30 & 82.98 & 81.68 \\
  ExpeL  &  83.93 & 72.19 & 69.52 & 85.36 & 81.53 & 82.31 \\
  Direct &  83.27 & 71.96 & 68.43 & 85.36 & 83.02 & 81.45 \\
  Init  &  84.19 & 72.38 & 69.58 & 85.42 & 82.55 & 81.81 \\
  SEO    & \textbf{84.23} & \textbf{72.61} & \textbf{70.11} & \textbf{85.47} & \textbf{83.05} & \textbf{82.40} \\
  \bottomrule
  \toprule
  Language & \multicolumn{3}{c|}{En-Zh} & \multicolumn{3}{c}{Zh-En} \\
  \hline
  Models & GPT-3.5 & Llama-2-13b & Llama-2-7b & GPT-3.5 & Llama-2-13b & Llama-2-7b \\
  \hline
  CLIN   &  86.37 & 71.79 & 72.56 & 81.13 & 77.37 & 76.50 \\
  ExpeL  &  86.41 & 73.18 & 72.48 & 81.07 & 77.38 & 76.46 \\
  Direct &  86.37 & \textbf{73.41} & 70.27 & 81.10 & 77.52 & 75.97 \\
  Init  &  86.41 & 66.81 & 72.59 & 81.07 & 77.46 & 76.40 \\
  SEO    & \textbf{86.50} & 73.23 & \textbf{73.05} & \textbf{81.17} & \textbf{77.62} & \textbf{76.67} \\
  \bottomrule
  \end{tabular}
  }
  \caption{COMET scores of various generator models for WMT23 En$\leftrightarrow$De and En$\leftrightarrow$Zh on both directions.\label{tab:mt}}
\end{table*}

On the contrary, by applying the experience obtained via SEO, we are able to consistently improve the performance of both direct answer without experiences and the answer with the initial experience across all three generator models. The results imply that SEO is capable of finding effective model-specific experiences.

\subsection{Machine Translation}

Table~\ref{tab:mt} shows the results for MT on En-De and En-Zh directions. Here, the improvement brought by SEO for GPT-3.5 is relatively slight. As the direct translation performance of GPT-3.5 is nearly state-of-the-art compared to the best systems of WMT23 \citep{Kocmi23findings}, it may be difficult for GPT-3.5 to largely boost the performance. However, for Llama models whose direct translation quality lags behind GPT-3.5, the performance of SEO experience almost consistently improves across four language directions. An exception case is in En-Zh where Llama-2-13b does not improve over Direct. This is mostly due to an unnatural performance drop of the initial experience that makes it too difficult to recover. Nevertheless, the SEO performance still excels Init and the other two baselines CLIN and ExpeL.

\begin{table}[!ht]
  \center
  \scalebox{0.95}{
  \begin{tabular}{l|ccc}
  \toprule
  Dataset & \multicolumn{3}{c}{CoLA} \\
  \hline
  Models & GPT-3.5 & Llama-2-13b & Llama-2-7b\\
  \hline
  CLIN  & 82.93 & 72.20 & 56.95 \\
  ExpeL &  83.22 & 58.58 & 65.58 \\
  Direct &  80.35 & 62.32 & 59.25 \\
  Init  &  81.02 & 64.91 & 59.92 \\
  SEO    & \textbf{83.60} & \textbf{74.11} & \textbf{72.77} \\
  \bottomrule
  \toprule
  Dataset & \multicolumn{3}{c}{SST-2} \\
  \hline
  Models & GPT-3.5 & Llama-2-13b & Llama-2-7b \\
  \hline
  CLIN  & 93.69 & 86.24 & 79.24 \\
  ExpeL & 93.23 & 87.72 & 83.83 \\ 
  Direct & 94.84 & 87.04 & 88.99 \\
  Init & 94.04 & 88.30 & 89.22 \\
  SEO   & \textbf{95.07} & \textbf{91.28} & \textbf{90.60} \\
  \bottomrule
  \end{tabular}
  }
  \caption{Accuracy scores of various generator models for CoLA and SST-2 tasks.\label{tab:classification}}
\end{table}

We also find that the improvements of translation into English are less significant compared to translation out of English. Since these LLMs are dominantly trained using English data \citep{Touvron23llama}, they possess stronger capability in generating English rather than other languages.

\subsection{Text Classification}

We also perform SEO for two text classification tasks. As shown in Table~\ref{tab:classification}, by applying the experiences obtained via SEO, all three models can boost their performance on both tasks. We observe that Llama models improves their performance by a large margin with SEO experience on CoLA. The results of direct answer for Llama models are much worse compared to GPT-3.5. This is possibly due to a mismatch of prompt used for the generator models. Note that we have not performed intense prompt engineering for each task. Instead, we use the same task prompt for all generator models. The results suggest that using SEO experience may also recover the performance degradation of imperfect prompts.

\section{Analyses}

\subsection{Effect of Initial Experience\label{ssec:init}}

As introduced in Section~\ref{sec:seo}, SEO requires an initial experience to launch the optimization process. In this section, we study the effect of different initialization. By default, we ask GPT-4 to generate an initial experience:
{\small
\begin{tcolorbox}[colback=gray!5!white,colframe=gray!75!black,title=Prompt for initial experience generation]
  Please list a few rules that could be useful for doing multi-hop question answering from several retrieved paragraphs. Rules should be concise and easy to follow.
\end{tcolorbox}
}
We only provide the task name to GPT-4 and do not include any specific task information. In addition, we reuse the ``human-crafted insights'' for HotpotQA released by \citet{Zhao23expel} as another initialization. These experiences are supposed to have a positive effect, as they are carefully designed to mitigate errors made by LLMs. Moreover, we also consider a third initialization where we ask GPT-4 to summarize experiences from some few-shot examples. We use the first five examples of HotpotQA annotated by \citet{Trivedi23interleaving} as few-shot examples.

\begin{table}[!ht]
  \center
  \scalebox{0.95}{
  \begin{tabular}{l|ccc}
  \toprule
  Models & Default & Human & From Few-shot \\
  \hline
  Direct & 37.2 & 37.2 & 37.2 \\
  Init  & 38.6 & 29.2 & 30.6 \\
  SEO    & \textbf{45.2} & \textbf{44.8} & \textbf{42.2} \\
  \bottomrule
  \end{tabular}
  }
  \caption{Performance of different initial experiences for GPT-3.5 on HotpotQA measured by EM score. \textit{From Few-shot} is the initial experience summarized from few-shot examples.\label{tab:score-init-mem}}
\end{table}

Table~\ref{tab:score-init-mem} illustrates the result of experiences optimized from different initial experiences using GPT-3.5 as the generator model. Without optimization, the impact of initial experiences exhibits considerable variability. Both human-crafted experience and experience summarized from few-shot examples has a negative effect. However, after performing SEO, both experiences can finally improve the performance. The optimized human-crafted experience delivers similar results to the default initialization. Among these initialization settings, the default initialization is simpler to conduct, as it does not require any human annotation or efforts. It also yields the best performance. Therefore, we use the default initialization for the following analyses.

\subsection{Impact of Experience}

We conduct further analyses to better understand how the experiences helps LLMs. For LLMs, it is not easy to always obtain the exact output in desired format, especially for multi-hop QA which uses EM as the metric. Therefore, it is unknown whether the performance improvement of SEO experience comes from generating answer in better format or really guiding LLMs in terms of correctness. To this end, we manually analyze the type of errors made by direct answer and using SEO experience on the dev set of HotpotQA. We classify errors into two types: format issue and wrong answer. The former indicates LLMs give a correct answer but does not match exactly the correct answer. We compute the differences between Direct and SEO for each category. As shown in Table~\ref{tab:qa-effect}, the performance improvement brought by SEO experience mainly comes from  generating more correct answers. In addition, it also guides LLMs to give answers in better format that suits better the EM score.

\begin{table}[!ht]
  \hspace{-4pt}
  \scalebox{0.95}{
  \begin{tabular}{l|ccc}
  \toprule
  Models & GPT-3.5 & Llama-2-13b & Llama-2-7b \\
  \hline
  \hspace{-6pt}$\Delta$Correct & 10 &  14 & \hphantom{0}6 \\
  \hspace{-6pt}$\Delta$Format  & -3 &  -7 & -1 \\
  \hspace{-6pt}$\Delta$Wrong   & -7 &  -7 & -5 \\
  \bottomrule
  \end{tabular}
  }
  \caption{Statistics of differences between SEO and Direct for correct answer, format issue and wrong answer on the dev set of HotpotQA for various LLMs. \label{tab:qa-effect}}
\end{table}

\begin{table}[!ht]
  \center
  \scalebox{0.95}{
  \begin{tabular}{l|cc|cc}
  \toprule
  Lang. & \multicolumn{2}{c|}{En-De} & \multicolumn{2}{c}{De-En} \\
  \hline
        & \# Fail & COMET$^*$ & \# Fail & COMET$^*$ \\
  \hline
  CLIN & \hphantom{0}\hphantom{0}9 & 70.06 & \hphantom{0}\textbf{0} & 81.68 \\
  ExpeL & \hphantom{0}\hphantom{0}\textbf{6} & 69.85 & \hphantom{0}\textbf{0} & 82.31 \\
  Direct & \hphantom{0}32 & 70.40 & \hphantom{0}8  & 81.84 \\ 
  Init & \hphantom{0}12 & 69.99 & \hphantom{0}\textbf{0} & 81.81 \\
  SEO    &  \hphantom{0}\hphantom{0}7 & \textbf{70.48} & \hphantom{0}\textbf{0}  & \textbf{82.40} \\
  \bottomrule
  \toprule
  Lang. & \multicolumn{2}{c|}{En-Zh} & \multicolumn{2}{c}{Zh-En} \\
  \hline
        & \# Fail & COMET$^*$ & \# Fail & COMET$^*$ \\
  \hline
  CLIN & \hphantom{0}11 & 72.70 & \hphantom{0}7 & 76.61 \\
  ExpeL & \hphantom{0}15 & 72.74 & \hphantom{0}\textbf{0} & 76.46 \\
  Direct & 143 & 72.75 & 22 & 76.26 \\
  Init & \hphantom{0}11 & 72.78 & \hphantom{0}\textbf{0} & 76.40 \\
  SEO    &  \hphantom{0}\hphantom{0}\textbf{8}  & \textbf{73.19} &  \hphantom{0}1 & \textbf{76.68} \\
  \bottomrule
  \end{tabular}
  }
  \caption{Number of failed translations (\# Fail) and COMET scores of Llama-2-7b model without taking failed translations into account (COMET$^*$) for MT.\label{tab:mt-effect}}
\end{table}

We also observed cases in Llama-2-7b models for MT tasks, for which the model fails to generate a translation of the input sentence. For most of the cases, Llama-2-7b refuses to translate and output irrelevant explanations due to security reasons. We analyze this situation and find that all experiences are able to greatly reduced the number of cases where Llama-2-7b fails to translate for all four directions, as illustrated in Table~\ref{tab:mt-effect} (\# Fail). Among all baselines, SEO experiences achieve the most significant reduction in general. Since translation failure has a negative effect on the overall COMET score, we further eliminate those cases and compute the COMET score on the remaining sentences that actually have a translation. Results in Table~\ref{tab:mt-effect} indicate that SEO experience still surpasses all other methods in terms of translation quality after removing translation failures. 

\subsection{Out-of-distribution Transfer}

When optimizing the experience, we guide the optimizer model to generate general experience that could be applied to more questions and do not contain question specific contents. We expect that the optimized experience could generalize not only to the same dataset, but also to other out-of-distribution data on the same task. To evaluate the generality of the optimized experience, we test it on 2WikiMultiHopQA \citep{Ho20constructing}, which is also a multi-hop QA dataset. Results in Table~\ref{tab:ood} show that both GPT-3.5 and Llama-2-13b can benefit from the experience optimized on HotpotQA data to directly improve the performance on 2WikiMultiHopQA. 

\begin{table}[!ht]
  \center
  \scalebox{0.95}{
  \begin{tabular}{l|cc}
  \toprule
  Models & GPT-3.5 & Llama-2-13b \\
  \hline
  Direct & 23.0 & 20.2 \\
  Init  & 24.4 & 15.4 \\
  SEO (HotpotQA)   & \textbf{35.4} & \textbf{27.0} \\
  \bottomrule
  \end{tabular}
  }
  \caption{EM scores of out-of-distribution results on 2WikiMultiHopQA using the experience optimized on HotpotQA.\label{tab:ood}}
\end{table}

\begin{figure}[ht]
  \center
  \includegraphics[width=\columnwidth]{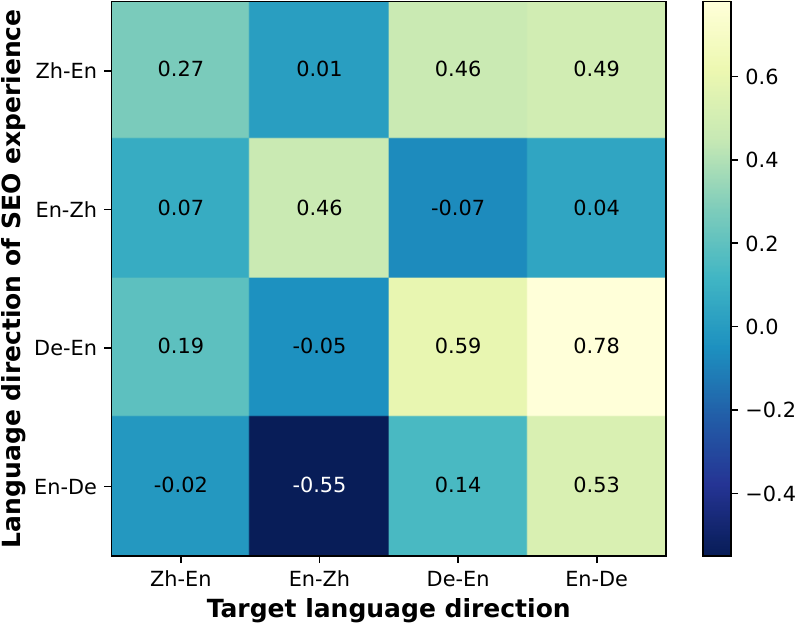}
  \caption{Differences of COMET scores when evaluating experience across language directions for Llama-2-7b. Scores in each column is calculated by subtracting the COMET score of initial experience for the corresponding direction. The diagonal reports improvement of applying experience on its own direction.\label{fig:heatmap}}
\end{figure}

\subsection{Transfer among Languages}

In the MT scenario, as illustrated in Figure~\ref{fig:opt-prompt}, we do not specify a translation direction in the prompt used for the optimizer model. Instead, we only indicate the task is MT. Therefore, we would like to evaluate the cross-lingual ability of optimized experience. To do so, we use the Llama-2-7b model and apply the experience optimized for one direction to the other three directions. We repeat this process for all four directions and obtain results in Figure~\ref{fig:heatmap}. We report the difference of COMET score between the tested results and results using the initial experience (Init), as we observe that the initial experience leads to improved performance compared to direct translation. In most cases, applying experience cross-lingually indeed enhances the translation quality over the initial experience. Surprisingly, for some directions like applying De-En experience to En-De, the performance is even better than using the experience optimized on its own direction. The results imply that experiences possesses the ability to generalize across languages.

\subsection{Ablation Study}

\begin{table}[!ht]
  \center
  \scalebox{0.95}{
  \begin{tabular}{l|cc}
  \toprule
  Models & Test & Dev\\
  \hline
  Direct & 37.2 & 41.0 \\
  Init  & 38.6 & 40.0 \\
  SEO    & \textbf{45.2} & \textbf{51.0} \\
  \quad GPT-3.5 as optimizer    & 44.6 & 49.0 \\
  \quad fix-valid                        & 44.4 & 47.0 \\
  \quad- validation                      & 37.6 & 44.0 \\
  \quad- two answers  ($\hat{Y}_{E_t}$ and $\hat{Y}$) & 40.2 & 48.0 \\
  \quad- exp effect ($\delta_{E_t}$)               & 42.6 & 49.0 \\
  \bottomrule
  \end{tabular}
  }
  \caption{EM scores when removing different components for GPT-3.5 on HotpotQA test and dev set. \textit{GPT-3.5 as optimizer} use GPT-3.5 instead of GPT-4 as the optimizer model. \textit{fix-valid} use a fixed validation set at each step. \textit{-validation} removes the validation step. \textit{-two answers} does not provide the response trials of the generator model to the optimizer model. \textit{-exp effect} omit the experience effect to the optimizer model.\label{tab:ablation}}
\end{table}

To better understand the effectiveness of each component in SEO, we conduct an ablation study. We use GPT-3.5 as the generator model and rerun SEO on HotpotQA by removing elements each time. We first use GPT-3.5 as the optimizer model. Result in Table~\ref{tab:ablation} shows that a weaker optimizer model lead to slightly worse performance, which is inline with \citep{Zhao23expel}. 

We also experiment SEO by using a fixed validation set instead of random sample at each iteration step during training. Table~\ref{tab:ablation} (fix-valid) indicates that using a fixed validation set may lead to less effective experience. 

Among all components, removing the experience validation step (-validation) results in the most significant performance drop. This indicates that the validation step plays the most important role in SEO to make sure that updated experience actually improves model performance. Moreover, removing the answers $\hat{Y}$ and $\hat{Y}_{E_t}$ without and with experience from the input to the optimizer model (-two answers) or removing the experience effect $\delta_{E_t}$ also leads to significant drop. As mentioned in Section~\ref{ssec:exp-update}, both information is necessary for the optimizer model produce better experiences.

We also observe that removing components from SEO tends to result in experiences that are less general. As shown in the column Dev of Table~\ref{tab:ablation}, the performance of best experiences on the dev set for ablation experiments does not drop too much. However, the test performance lags far behind. The results suggest that the input elements to the optimizer model tends to help in increasing the generality of optimized experience.

\begin{table*}[!ht]
  \center
  \scalebox{0.99}{
  \small
  \begin{tabularx}{\linewidth}{X}
  \toprule
  1. Preprocessing: Use a robust retrieval method to retrieve paragraphs that may contain relevant information. Discard paragraphs that are irrelevant or contain too many noise. Narrow down the retrieved paragraphs to a manageable size. \\
  2. Paragraph Analysis: Analyze the retrieved paragraphs individually to identify the most relevant passages. Focus on key concepts, facts, and arguments mentioned in each paragraph. Ignore redundant or unrelated information. \\
  3. Entity Extraction: Extract named entities (such as people, organizations, locations) from the retrieved paragraphs. These entities can provide important clues for answering multi-hop questions by connecting information across paragraphs. \\
  \textcolor{blue}{4. Co-reference Resolution: Resolve pronouns or ambiguous references within and across paragraphs. This ensures that the various mentions of the same entity are correctly linked together, enabling coherent understanding of the retrieved information.} \\
  \textcolor{blue}{5. Relation Extraction: Identify relationships between entities mentioned in different paragraphs. Look for dependencies, associations, or connections between the extracted entities to answer the multi-hop questions effectively.} \\
  \textcolor{red}{6. Hierarchical Structure: Consider the hierarchical structure of the retrieved information. Some paragraphs may provide general context, while others may contain more specific details or evidence. Utilize this structure to determine the flow of information and prioritize the relevant paragraphs accordingly.} \\
  7. Logical Reasoning: Apply logical reasoning to connect pieces of information across multiple paragraphs. Inferential reasoning, deduction, or induction can help answer multi-hop questions that require drawing conclusions from scattered information. \\
  \textcolor{red}{8. Confidence Scoring: Assign confidence scores to each retrieved paragraph based on relevance and reliability. Consider factors such as the credibility of the source, the coherence of the content, and the consistency of the information mentioned. Give more weight to paragraphs with higher confidence scores during the answer generation process.} \\
  \textcolor{blue}{9. Answer Generation: Combine the relevant information extracted from multiple paragraphs to form a coherent answer. Pay attention to the question constraints, and try to generate an answer that is concise, accurate, and supported by the retrieved information.} \\
  10. Iterative Approach: If an initial answer based on one set of paragraphs is not satisfactory, employ an iterative approach. Revise the retrieval process, reanalyze the retrieved paragraphs, or refine the answer generation to improve the results. Iterate until a satisfactory answer is achieved. \\
  \bottomrule
  \end{tabularx}
  }
  \caption{Initial experience for multi-hop QA. Unhelpful sentences that are further removed are marked in red. Useful sentences that are almost kept are in blue.\label{tab:init-mem}}
\end{table*}

\begin{table*}[!ht]
  \center
  \scalebox{0.99}{
  \small
  \begin{tabularx}{\linewidth}{X}
  \toprule
  1. Relevance Determination: Discard paragraphs that do not contain information relevant to the question. Retain those that mention the specific entities, keywords, or time periods from the question. Prioritize paragraphs containing both entities in the question. \\
  2. Keyword and Phrase Extraction: Extract key phrases and keywords from the relevant paragraphs. These include named entities, time periods, and specific terms connected to the question. \\
  3. Key Fact Identification: Identify key facts and details in the relevant paragraphs. Look for information that directly answers the question, such as actions, events, or results related to the entities in the question. \\
  \textcolor{blue}{4. Context Understanding: Understand the context of the keywords and key facts. Analyze the sentences and paragraphs where they appear to comprehend their specific meaning in relation to the question.} \\
  5. Co-reference Resolution: Resolve pronouns or ambiguous references within and across paragraphs. Ensure that all mentions of the same entity are correctly linked together. \\
  6. Relation Extraction: Identify relationships between entities mentioned in the paragraphs. Look for connections between the extracted entities that could directly answer the question. \\
  \textcolor{blue}{7. Semantic Analysis: Understand the semantic structure of the question. Identify what the question is asking for (e.g., a specific type, a relationship, a comparison, a cause or effect, etc.). This helps in targeting the information extraction and synthesis process.} \\
  8. Information Synthesis: Combine information from multiple relevant paragraphs to infer the answer. This could involve connecting details about the same entity from different paragraphs. Prioritize information that directly addresses the question. \\
  9. Answer Generation: Use the relevant information extracted from the paragraphs to form a coherent answer. The answer should be concise, accurate, and supported by the retrieved information. Ensure that the answer correctly reflects the specific entities or details asked in the question. \\
  10. Answer Verification: Cross-check the generated answer with the retrieved paragraphs to ensure its accuracy. If the initial answer is not satisfactory, revisit the paragraphs, reanalyze the information, or refine the answer generation process until a satisfactory answer is achieved. \\
  \textcolor{blue}{11. Answer Formatting: Ensure the answer is formatted to match the expected answer format. For example, if the question asks for a yes or no response, the answer should be formatted as "yes" or "no" rather than in a sentence form.} \\
  \bottomrule
  \end{tabularx}
  }
  \caption{Experience optimized by SEO for GPT-3.5 on HotpotQA. Emergent sentences are marked in blue.\label{tab:best-mem}}
\end{table*}

\subsection{Experience Evolution}

We here perform a qualitative study to show the change from the initial memory to the optimized one. We present the initial experience for multi-hop QA and the corresponding optimized experience for GPT-3.5 in Tables~\ref{tab:init-mem} and \ref{tab:best-mem}, respectively. The initial experience contains rules like ``Hierarchical Structure'' or ``Confidence Scoring'' (marked in red) that may not be helpful, as they are completely removed by the optimizer model. Rules like ``Co-reference Resolution'', ``Relation Extraction'' and ``Answer Generation'' are likely to be useful. They mostly remain in the optimized experience. However, there are still some minor paraphrasing of these rules. We conjecture that these paraphrasing are for the purpose to suit specific generator model. 

In the optimized experience, there are also some emergent rules that appear to be very useful from a human perspective. For example, the ``Semantic Analysis'' rule mentions understanding what the question is asking for, which we think is important for correctly answer the questions. The ``Answers Formatting'' rule points out that the answer should be in a correct format, which directly helps in the computation of EM score.

We also show an example that applying SEO experience helps generate the correct answer in Appendix~\ref{sec:case}.

\section{Related Work\label{sec:related}}

\paragraph{Using Experiences in LLMs}

\citet{Dalvi22towards} perform question answering by interacting with humans and use human generated feedback as experience to help models perform better in answering questions. \citet{Madaan22memory,Yang23failures} intend to use LLM instead of human to generate answer feedback. \citet{Shinn23reflexion} propose the self-reflection approach that ask LLMs to reflect over its own failure trajectory when performing a specific task case and generate reflections that may help the model to avoid failures. The reflection is added in the next try for the same case to improve the success rate. \citet{Madaan23selfrefine} share the similar idea by asking LLM to generate feedback about its previous answer and using feedback to refine its answer for the same case. Feedbacks generated by the above approaches are often specific to certain cases that require retrieval based input question or only reusable to the same question. 
In this paper, we tend to optimize experiences by SEO that generally improves performance on a type of task, rather than on certain cases.

Some studies have also explored to generate general experiences for LLMs. \citet{Zhao23expel} ask LLMs to compare between success and failure self-reflection trajectories to iteratively generate general experiences. They use a scoring mechanism to sort each generated rules and remove rules when they score 0. \citet{Majumder23clin} also reflect on LLM-generated trials to iteratively update experiences. They restrict the format of each rule to be causal abstractions between an action and a result. Our SEO approach, on the contrary, does not require specific design. Instead, it contains an experience validation step to verify that updated experience is actually helpful for the same task, therefore avoiding unavailing updates. Moreover, SEO is capable of generating model specific experiences that steadily improves different LLMs.

\paragraph{Automatic Prompt Optimization using LLMs}

Several works have studied to automatically optimize the prompt used to perform specific tasks in order to obtain better performance. \citet{Zhou23human} use LLMs to summarize task instructions from input-output example pairs and select best performing prompt. \citet{Pryzant23automatic} ask LLMs to reflect over a prompt, consider the generated flow as the gradient, and edit the prompt in the opposite semantic direction of the gradient. They also propose to keep multiple candidate prompts during each iteration via a beam-search like strategy. \citet{Yang24optimizers} provide multiple historical prompts and the corresponding performance score to the optimizer model to explore new prompt that leads to better scores. \citet{Tang24unleashing} further propose to limit the degree of prompt modification in each optimization step by computing the edit distance between old and new prompt to ensure gradual changes of prompt. Prompt optimization methods focus on finding task prompts that aim to give instructions to LLMs to perform a specific task. However, in SEO, we optimize a different objective. We keep the same task prompt and optimize experiences which provide guidance about how to perform a specific task.

\section{Conclusion}

In this work, we proposed SEO, an experience optimization framework, that was able to find useful and model-specific experience that improves the performance of various LLMs on specific tasks without updating their parameters. SEO iteratively updates experience and validate the experience on a stochastic validation set to ensure the effectiveness of updated experience at each training step. Experiments on seven datasets across three different tasks using three LLMs demonstrated that SEO was generally stable in producing optimized model-specific experiences. We also found that experiences obtained by SEO not only improved the performance of LLMs on the same dataset, but also generalized to out-of-distribution data on similar tasks.

\section*{Acknowledgement}

We would like to thank professor Yang Liu and Peng Li for their great help and profound suggestions for this work. Jitao Xu is supported by the China Postdoctoral Science Foundation under Grant Number 2023TQ0245.

\bibliography{biblio}
\bibliographystyle{tacl_natbib}

\clearpage
\appendix 
\onecolumn

\section{Case Example\label{sec:case}}

Table~\ref{tab:example} illustrates an example that applying SEO experience helps generate the correct answer.

\begin{table*}[h]
  \center
  \scalebox{0.99}{
  \begin{tabularx}{\linewidth}{l|X}
  \toprule
  Q & Which has more members, Dada or Alt-J? \\
  \hline
  Ref & Alt-J \\
  \hline
  Direct & So the answer is: Dada \\
  \hline
  SEO & To answer the question, we need to determine the number of members in both Dada and Alt-J. Let's analyze the relevant paragraphs:

1. Paragraph: Dada (band)

   $\quad$- Mentions Dada as a three-piece rock band from California.
   
   $\quad$- Provides the names of the band members: Michael Gurley, Joie Calio, and Phil Leavitt.

2. Paragraph: Alt-J

   $\quad$- Mentions Alt-J as an English indie rock band formed in 2007 in Leeds.
   
   $\quad$- Provides the names of the band members: Gwil Sainsbury, Joe Newman, Thom Sonny Green, and Gus Unger-Hamilton.

Based on the information provided, Dada has 3 members, while Alt-J has 4 members. Therefore, Alt-J has more members than Dada.

So the answer is: Alt-J. \\
  \bottomrule
  \end{tabularx}
  }
  \caption{Example answer using optimized experience for GPT-3.5 on HotpotQA to obtain correct answer compared to direct answer. We ask LLMs to generate the term ``So the answer is:'' that is used to parse and collect final answer.
\label{tab:example}}
\end{table*}

\section{Additional Prompts\label{sec:prompt}}

We show prompts used for the generator models on different tasks in the following.

\begin{figure*}[h]
  \center
\begin{tcolorbox}[colback=gray!5!white,colframe=gray!75!black,title=Prompt for multi-hop QA with GPT-3.5]
You are an advanced reasoning agent that can solve question answering tasks by using information from given paragraphs and reasoning step by step. Answer the following question. After reasoning steps, give your answer in the format: "So the answer is: <ANSWER>".
\\
\\
Through multiple rounds of search, you have obtained the following paragraphs:
\\
\\
<Info>
\\
\{\texttt{context}\}
\\
</Info>
\\
\\
These paragraphs may contain useful information that could help answer the question. There is also information that are not helpful.
\\
\\
Question: \{\texttt{question}\}
\\
\\
Answer:
\end{tcolorbox}
\end{figure*}

\begin{figure*}[ht]
  \center
\begin{tcolorbox}[colback=gray!5!white,colframe=gray!75!black,title=Prompt for multi-hop QA with Llama-2 models]
You are an advanced reasoning agent that can solve question answering tasks by using information from given paragraphs and reasoning step by step. Answer the following question. Give your reasoning steps, followed by your answer in the format: "So the answer is: <ANSWER>". The final answer should only contain the answer span.
\\
\\
Retrieved paragraphs:
\\
\\
<Info>\\
\{\texttt{context}\}\\
</Info>
\\
\\
These paragraphs may contain useful information that could help answer the question. There is also information that are not helpful.
\\
\\
Question: \{\texttt{question}\}
\\
\\
Answer: 
\end{tcolorbox}
\end{figure*}

\begin{figure*}[ht]
  \center
\begin{tcolorbox}[colback=gray!5!white,colframe=gray!75!black,title=Prompt for MT]
You are an excellent machine translation system. Translate the following input sentence from \{\texttt{src}\} to \{\texttt{tgt}\}. Give your translation in the format: "The translation is: <TRANSLATION>". The final translation should only contain the translated sentence.
\\
\\
Input: \{\texttt{sentence}\}
\\
\\
Translation: \\
\end{tcolorbox}
\end{figure*}

\begin{figure*}[ht]
  \center
\begin{tcolorbox}[colback=gray!5!white,colframe=gray!75!black,title=Prompt for CoLA]
You are an expert classifier. You are now tasked with handling a text classification task, which determines the grammatical correctness of a given sentence. If the sentence is grammatically correct, give a label of 1; otherwise, give a label of 0. Give your label results as "The label is: 0" or "The label is: 1".
\\
\\
Input sentence: \{\texttt{sentence}\}
\\
\\
The label is:\\
\end{tcolorbox}
\end{figure*}

\begin{figure*}[ht]
  \center
\begin{tcolorbox}[colback=gray!5!white,colframe=gray!75!black,title=Prompt for SST-2]
You are an expert classifier. You are now tasked with handling a text classification task, which determines the sentiment of a given sentence. If the sentence has a positive sentiment, give a label of 1; otherwise, give a label of 0. Give your label results as "The label is: 0" or "The label is: 1".
\\
\\
Input sentence: \{\texttt{sentence}\}
\\
\\
The label is:\\
\end{tcolorbox}
\end{figure*}

\end{document}